\documentclass{article}
\usepackage[utf8]{inputenc}
\usepackage[a4paper, width=155mm, top=35mm, bottom=35mm]{geometry}
\usepackage{graphicx}

\usepackage{booktabs}
\usepackage{style}
\usepackage{caption}
\usepackage{subcaption}

\usepackage{amsmath}

\def\TheTitle{\textbf{How User Language Affects Conflict Fatality Estimates in ChatGPT %/ Language Bias in ChatGPT Conflict Fatality Estimates
}}
\title{\Large \vspace{-3cm} \begin{onehalfspace} \TheTitle \end{onehalfspace} \vspace{3ex}}
\author{
  \large Daniel Kazenwadel$^{\dagger,1}$
\and
  \large Christoph Valentin Steinert$^{\dagger,2}$}
\date{ \vspace{2ex} \large \today}

\def\TheAbstract{OpenAI's ChatGPT language model has gained popularity as a powerful tool for complex problem-solving and information retrieval. However, 
concerns arise about the reproduction of biases present in the language-specific training data. In this study, we address this issue in the context of the Israeli-Palestinian and Turkish-Kurdish conflicts. Using GPT-3.5, we employed an automated query procedure to inquire about casualties in specific airstrikes, in both Hebrew and Arabic for the former conflict and Turkish and Kurdish for the latter. Our analysis reveals that GPT-3.5 provides 27$\pm$11 percent lower fatality estimates when queried in the language of the attacker than in the language of the targeted group. % Qualitative evidence further support these findings. 
Evasive answers denying the existence of such attacks further increase the discrepancy, creating a novel bias mechanism not present in regular search engines. This %highlights GPT-3.5's substantial 
language bias %, which 
has the potential to amplify existing media biases and contribute to information bubbles, ultimately reinforcing conflicts.
}

\title{How User Language Affects Conflict Fatality Estimates in ChatGPT}
\author{Daniel Kazenwadel$^{\dagger,1}$ \& Christoph V. Steinert$^{\dagger,2}$}
\begin{document} 

\thispagestyle{empty} \enlargethispage{1cm}
\noindent \maketitle 

%\hspace{4.2cm} \textit{Working paper in progress.} \\[2ex]

\begin{onehalfspace}
\vspace{0.5cm} \noindent \TheAbstract
\end{onehalfspace}

\vfill{}

\begin{singlespace}
\begin{footnotesize} \noindent
$^\dagger$ both authors contributed equally\\[1ex]
$^1$University of Konstanz, \texttt{\href{mailto:daniel.kazenwadel@uni-konstanz.de}{daniel.kazenwadel@uni-konstanz.de}}  \\[1ex]
$^2$University of St. Gallen, \texttt{\href{mailto:christoph.steinert@unisg.c}{christoph.steinert@unisg.ch}}  \\[2ex]

%Acknowledgements here 
\end{footnotesize}
\end{singlespace} \vspace{2ex}

\pagebreak

% %%%%%%% Anonymous first page %%%%%%%%
% %\begin{comment}
% \thispagestyle{plain}

% \begin{center}
% \noindent {\Large \TheTitle }
% \end{center}

% \begin{onehalfspace}
% \vspace{4cm} \noindent
% \TheAbstract 
% \end{onehalfspace}

% \pagebreak
% %\end{comment}

\section{Introduction}

Scholars have long recognized that information discrepancies play a profound role in armed conflicts \citep{fearon_rationalist_1995,slantchev_mutual_2011}. %Examples of wars that have been fueled by divergent beliefs include but are not limited to the Spanish-American War, the First World War, the Iraq War, or Russia’s war of aggression in Ukraine. 
Discrepancies in information have affected armed conflicts throughout history, but what distinguishes today's conflicts is the availability of an unprecedented amount of information sources. %Individuals can nowadays draw on abundant information from the internet, and it is plausible that artificial intelligence (AI) will increasingly be used as a powerful new information source.
Nowadays, individuals can draw on abundant online information about conflict-related events and even employ artificial intelligence (AI) to obtain targeted answers to specific questions.\footnote{This may not apply to citizens of states where the Internet is completely censored. However, there are only a small number of states where internet censorship is comprehensive, and individuals tend to be savvy in circumventing these restrictions \citep[e.g.,][p. 328]{king_how_2013}.} To the extent that these new sources of information mitigate information discrepancies and contribute to a convergence of beliefs, they may have a pacifying effect on conflict-prone regions.

However, quite to the contrary, it has been argued that these novel technologies facilitate the spread of misinformation and reinforce radical beliefs \citep{koehler_radical_2014,zhuravskaya_political_2020}. As people tend to search for belief-congruent information, targeted algorithms can create `information bubbles’ that reproduce prior beliefs \citep{kaakinen_shared_2020,rhodes_filter_2022}. Being confronted with fake news and deepfakes, it may be harder than ever before to identify the correct information. This is especially true when the novel information sources themselves have built-in biases that affect the content of the information obtained. While AI provides an appearance of objectivity, the information obtained may differ between people who speak different languages. For instance, the popular chatbot ChatGPT relies on the logic of prompting, meaning that the answers obtained are a function of the information provided in the question prompt. In multilingual contexts, individuals are likely to provide question prompts in different languages which may shape the content produced by the language-based model. How this affects information discrepancies in the context of multilingual armed conflict has not yet been investigated.  

Against this backdrop, this study investigates how language affects information about conflict-related violence obtained by GPT-3.5, the language model underlying ChatGPT. We analyze airstrikes that occurred during the Israeli-Palestinian conflict and the Turkish-Kurdish conflict as recorded by the UCDP Georeferenced Events Dataset \citep{davies_organized_2022, sundberg_introducing_2013}. For each airstrike, we ask GPT-3.5 for the number of people killed in both Hebrew and Arabic, and in Turkish and Kurdish respectively.\footnote{We use Northern Kurdish, also known as Kurmanji, which is the most widely spoken form of Kurdish. Kurmanji is predominantly used in south-eastern Turkey and northern Iraq, which are the regions covered in our paper.} %, keeping the question wording constant. 
By automating this process and repeatedly asking the same question about each airstrike, we obtain varying fatality numbers, which allow us to generate uncertainty estimates. Drawing on this quantitative information, we analyze how the language provided to GPT-3.5 affects the information obtained on airstrike fatalities. %We further leverage the content of GPT-3.5’s responses and apply sentiment analyses to explain cross-language differences in information on conflict-related violence.  

Our findings show for the first time that there is a substantial language bias in the information on conflict-related violence provided by GPT-3.5. The evidence suggests that GPT-3.5 reports higher casualty figures when asked about airstrikes in the language of the targeted group than in the language of the perpetrator.\footnote{We do not take a political stance on the question of who is the aggressor in the overall conflicts but use the terms `perpetrator' and `attacker' only in relation to specific airstrikes carried out by one side.} More specifically, in the context of Turkish airstrikes against alleged PKK members, we find that GPT-3.5 reports higher fatality numbers when it is asked about these airstrikes on Kurdish compared to Turkish. Similarly, we find that GPT-3.5 reports higher fatality numbers on Arabic compared to Hebrew in response to question prompts about Israeli airstrikes. Further we identify a new, previously unreported %, up to now unreported 
bias mechanism that exists only in chatbot responses and has no equivalent in traditional search engines. When asked in the language of the attacker, the chatbot not only provides a lower number of casualties but is also more likely to deny the existence of the queried event or reports an attack by the opposing side. Overall, GPT-3.5 is more likely to provide information about airstrikes and tends to produce higher casualty estimates in the language of the targeted group.

This evidence contributes to our understanding of political biases in AI \citep[][]{hartmann_political_2023,mcgee_is_2023} with a specific focus on fatality estimates in armed conflicts. While previous research demonstrates that AI is prone to gender biases \citep{leavy_gender_2018,marinucci_exposing_2022,nadeem_gender_2020} and racial biases \citep{cheng_marked_2023,intahchomphoo_artificial_2020,noseworthy_assessing_2020,obermeyer_artificial_2021,lee_detecting_2018}, we identify a novel language bias that shapes information discrepancies in multilingual conflicts. This speaks to previous research suggesting that intrastate conflicts occur more frequently within linguistic dyads than religious dyads \citep{bormann_language_2017}. By demonstrating that individuals are exposed to different information environments depending on their spoken language, we identify one mechanism linking multilingual contexts to conflict. More broadly, the evidence contributes to research on misinformation and propaganda during armed conflicts \citep{greenhill_rumor_2017,honig_evidence-fabricating_2018,lewandowsky_misinformation_2013,silverman_seeing_2021,schon_how_2021}. We show that new information technologies do not solve these problems, but reproduce biases that are widespread in media coverage of conflict-related violence.

As a major methodological contribution, we provide a novel tool for analyzing such language biases in large language models. We use the inherent translation capabilities of ChatGPT to translate and back-translate our prompts in a fully automated query scheme. This approach allows for good scalability and applicability to diverse topics and languages. Our focus on numerical estimates allows for statistical analysis and is not dependent on the subtleties of the exact translation and wording that affect more classical approaches such as sentiment analysis \citep[e.g.,][]{mohammad_challenges_2017}.

Our study proceeds as follows: First, we review previous research on deliberate misinformation campaigns and reporting biases during armed conflicts. Building on this body of research, we develop our theoretical expectations about the impact of language %differences 
on conflict-related information obtained through AI. Subsequently, we introduce our research design that enables us to test our hypothesis using an automated  procedure employed in GPT-3.5. Next, we present our results on estimates of airstrike fatalities in the Israeli-Palestinian conflict and in the Turkish-Kurdish conflict obtained through repeated multilingual searches in GPT-3.5. In the final section, we review these findings and draw broader implications on the nexus of AI and armed conflicts.

\begin{singlespace}
\section{Conflict fatality estimates, reporting biases, and novel sources of information}
\end{singlespace}

\noindent Information on conflict-related violence is a highly contested good. Belligerents have incentives to deny or inflate information about conflict-related violence given that battlefield objectives must be balanced against other concerns such as legitimacy \citep{malthaner_violence_2015,podder_understanding_2017,schlichte_armed_2015}, audience costs \citep{kurizaki_detecting_2015,slantchev_politicians_2006}, or combat morale \citep{fennell_morale_2014,nilsson_primary_2018}. Perpetrators of violence might want to downplay the extent of violent attacks to avoid negative repercussions such as domestic opposition or international sanctions. Evidence suggests that violence can trigger a backlash and incite opposition against the perpetrator \citep{carey_dynamic_2006,curtice_street-level_2021,rozenas_mass_2019,steinert_political_2022}. This is especially likely when violence is clearly attributable to one side \citep{thomson_grievance_2017} and when it is indiscriminate and causes civilian causalities---such as in the case of airstrikes \citep{pechenkina_how_2019,rozenas_mass_2019,schutte_how_2022}. In anticipation of possible adverse consequences, perpetrators of violence may seek to deny acts of violence or downplay their scale and intensity.\footnote{In some contexts, perpetrators of violence may wish to exaggerate the scale of their attacks in order to spread fear and signal their resolve. Such exaggeration of violence is particularly common in the case of terrorist groups that seek to maximise fear and attention \citep{blankenship_when_2018,braithwaite_logic_2013,pape_strategic_2003}.} All else equal, governments are in a privileged position to distort information about conflict-related violence because they can use state-controlled media and their own propaganda apparatus to whitewash their public image \citep{guriev_informational_2019}. However, evidence suggests that non-state actors also go to great lengths to portray themselves as norm-abiding actors, seeking to attract legitimacy and international support \citep{huang_rebel_2016,salehyan_explaining_2011,stanton_rebel_2020}.

On the other hand, victimized groups have incentives to inflate information on the scale of violence perpetrated by their opponent. By reporting (exaggerated) numbers of deaths caused by their opponent's attacks, they may seek to attract international solidarity and damage their opponent's reputation \citep{honig_evidence-fabricating_2018,noor_when_2012,silverman_seeing_2021}. In particular, reports of civilian casualties including allegations of attacks on vulnerable groups such as children, can be used strategically to portray the opponent as cruel and inhumane. Evidence further suggests that reports of female casualties attract stronger opposition than reports of male casualties \citep{kreft_imperfect_2023}. In order to appeal to international norms and reduce support for the perpetrators of violence, victimized groups tend to emphasize the indiscriminate and disproportionate nature of the violence perpetrated by their adversaries. In sum, belligerents are engaged in an information war for ``the hearts and minds'' of domestic and international audiences, resulting in strategic attempts to manipulate information about conflict-related violence.

While belligerents deliberately manipulate information, even independent sources may not be able to provide accurate information on conflict-related violence. Verifying information in war contexts is inherently difficult, given the imminent risk of violence and a disrupted information infrastructure \citep{nohrstedt_new_2014,saul_international_2008}. Physical obstacles such as blocked roads, destroyed bridges, and damaged power grids hamper the work of journalists and human rights organizations \citep{pfeifle_detecting_2022}. 
Fact-finding needs to be constantly adapted to local security concerns, as a significant number of journalists are killed while reporting in conflict societies \citep[see][p. 163]{gohdes_canaries_2017}. Because information is chronically difficult to verify, media reports of conflict-related violence tend to under-report the true incidence of violent events \citep{price_selection_2015,price_updated_2014}. This underreporting bias follows systematic patterns, as for example conflict-related violence in rural areas is less likely to be reported \citep{kalyvas_urban_2004}.

%Overall, citizens in conflict-affected countries find themselves in a complex information environment where it is difficult to obtain accurate evidence. Being confronted with sparse and often conflicting news, beliefs on conflict-related violence tend to be shaped by regime propaganda, narratives, and rumors \citep{schon_how_2021}. Political affiliations may influence which type of information source is deemed as credible. According to research on motivated reasoning, people search %disproportionately 
%for belief-congruent information to reinforce their priors \citep{nyhan_taking_2020}. However, it is plausible that motivated reasoning coexists with a genuine interest in the truth at least among a subset of the population. Evidence suggests that individuals have a strong desire to identify the truth on conflict-related violence as truth-seeking is linked to concerns for justice and dignity of the victims \citep{isaacs_at_2010,kim_massacres_2014,loyle_rebel_2021}. Hence, we expect that individuals in conflict-affected countries actively reach out for information on conflict-related violence. 

Overall, citizens in conflict-affected countries find themselves in a complex information environment where it is difficult to obtain accurate evidence. Novel information technologies facilitate access to information about conflict-related events but they can also reinforce political biases. %While the internet offers an unprecedented amount of information, it remains challenging to differentiate high-quality information from propaganda and fake news. 
Substantial evidence suggests that social media is prone to creating `information bubbles', fostering ideological polarization and radicalized identities \citep{dobransky_piercing_2021,eady_how_2019,kaakinen_shared_2020,spohr_fake_2017}. Chatbots such as ChatGPT offer a new source of information that can provide concise answers to specific questions. 
% Being currently primarily used by English-speaking tech enthusiast and content creators, it is plausible to extrapolate that AI will increasingly be used for information purposes among larger audiences. 
Being already used in diverse fields such as content creation \citep[][]{lawrence_can_2023}, programming \citep[][]{surameery_use_2023}, education \citep[][]{lo_what_2023}, research \citep[][]{zhu_chatgpt_2023}, tourism \citep[][]{carvalho_chatgpt_2023}, and marketing \citep[][]{rivas_marketing_2023}, it is plausible to extrapolate that large language models will increasingly be used for information purposes among larger audiences. In particular, they could be used to obtain information on complex and controversial issues---such as conflict-related violence---where it is difficult to find clear-cut information elsewhere.\footnote{The responses from ChatGPT do not reflect the level of uncertainty associated with the information. They therefore suggest a level of confidence that other sources may not offer.} 

In light of this ongoing development, it is important to understand how AI responds to questions about conflict-related violence. To date, we lack systematic empirical evidence on AI in this specific context. While AI may increasingly reach global audiences, we expect that individuals' engagement with AI will vary systematically across cultural contexts. In particular, we argue that language competence is a fundamental constraint on individuals' engagement with AI. Despite the fact that chatbots such as ChatGPT can be used as translators, for mere convenience purposes, it is plausible that individuals will primarily engage with AI through their own spoken and written language. 
It is well known that the quality of ChatGPT's answers depends strongly on the language used, since it relies mostly on the subset of the training data that is in the same language as the question \citep[][]{blevins_language_2022}. The training data---a mixture of different sources, including a copy of open access internet data (\href{https://commoncrawl.org/}{Common Crawl}), an overview of open source books, %(books1) XXX TODO: book 1 is not really informative, do you want us to quote this book??
and Wikipedia---is heavily biased towards the English language (over 50\%) %and the percentage of training data in each language is one-sided with the majority of over 50 percent being in English
\citep{artetxe_does_2022,crawl_archives_statistics_2023,dodge_documenting_2021}.
Since the performance of language models depends on the amount of training data, this results in a significantly worse performance for languages with less training data \citep{fang_how_2023,kreutzer_quality_2022,lai_chatgpt_2023}. %Using quantitative and qualitative methods We show here for the first time that besides the already established performance difference there is a strong difference in cultural values of ChatGPT resulting in different numbers when the model is asked to give quantitative answers. 
This means that even if people ask exactly the same question in different languages, the language model is expected to produce different answers. %, which reflect the cultural bias of the respective languages and therefore might amplify already existing biases and filter bubbles in that respective language, even when the model gives the impression of perfect objectivity. 

We argue that this has implications for queries on conflict-related violence. We expect that the information available in the training data on events of conflict-related violence will differ systematically across languages. In light of the incentives for information distortion discussed above, we hypothesize that ChatGPT responses differ depending on the language of the query. In particular, in the context of airstrikes during armed conflicts, we expect fewer reported deaths in the language of the attacker than in the language of the targeted group. %Hence, we hypothesize that conflict fatality estimates provided by ChatGPT are affected by a language bias.
\vspace{6mm}

%, leading us to derive the following hypothesis:

%\vspace{4mm}

%\begin{center} $H_{1}$: Conflict fatality estimates by GPT-3.5 are affected by language biases. \end{center}

\section{Research Design}

We investigate the hypothesis in the context of airstrikes during armed conflicts where the parties to the conflict are linguistically divided.\footnote{This excludes, among others, Russia's ongoing war of aggression in Ukraine as Russian is spoken in both countries.} We select the two cases of the Israeli-Palestinian conflict (Hebrew/Arabic dyad) and the Turkish-Kurdish conflict (Turkish/Kurdish dyad), which are both classified as intra-state armed conflicts by the UCDP/PRIO Armed Conflict Dataset \citep{gleditsch_armed_2002,davies_organized_2022}. These conflicts are comparable in the sense that professional armies are pitted against weaker non-state insurgents, representing typical cases of irregular warfare \citep{kalyvas_international_2010}. While holding the type of conflict constant, the analyzed conflicts differ substantially in historical, political, and cultural dimensions, allowing us to analyze whether our hypothesis holds in different multilingual contexts. We focus on airstrikes in these conflicts because this type of conflict-related violence tends to result in fatalities, but the numbers are often disputed and difficult to verify.\footnote{As only the Israeli and the Turkish governments have air forces, airstrikes can easily be attributed to the respective conflict parties.} The large number of conflicts in which English-speaking countries were involved (Vietnam, Afghanistan, Iraq, etc.) are excluded due to the significantly better performance of AI in this language  \citep[see][]{dodge_documenting_2021}. %TODO https://dl.acm.org/doi/10.1145/3597307. 

To identify airstrikes during the two conflicts under scrutiny, we use information from the  UCDP Georeferenced Event Dataset (GED), which contains fine-grained information on individual events of organized violence that are geo-coded to the level of individual villages \citep{davies_organized_2022,sundberg_introducing_2013}. The major advantage of this dataset is that it provides us with rich contextual information about individual airstrikes, such as the exact day and location, which allows us to pinpoint these airstrikes through specific questions in GPT-3.5. %Further, GED contains information on `official' fatality estimates, including upper and lower bounds, as reported by `independent' news sources, providing us with a benchmark for the fatality estimates reported by GPT-3.5.\footnote{Note that such official fatality estimates are also prone to biases in light of the inherent difficulties to obtain accurate information from conflict-torn societies \citep{eck_data_2012,gohdes_first_2013,spagat_estimating_2009}.} 
While GED covers different types of organized violent events, we filter the subset of airstrikes by searching for this term via string detection in the `source article' column provided by GED. Subsequently, we randomly select 10 airstrikes for both the Turkish-Kurdish conflict and the Israeli-Palestinian conflict, identified by the `conflict name' column in GED. All analyzed airstrikes were carried out by the Turkish government against Kurdish individuals or by the Israeli government against Palestinian individuals. We present the exact locations of these airstrikes on the map in panel (a) of the figures \ref{fig:arab_hebrew} and \ref{fig:kurman_turk}, based on the geolocation coordinates provided by GED.

Drawing on the information provided by GED, we then developed 10 short question prompts for each conflict to ask GPT-3.5 about the number of fatalities in each of these airstrikes. The questions include information about the perpetrator of an airstrike, the exact date of the airstrike, and the location where the airstrike took place. As an example of an Israeli airstrike, we used the question \textit{``In the Israeli airstrike on August 21, 2014 in the Nuseirat refugee camp how many were killed?''}.\footnote{We had tried different question wordings and chose this question based on the highest propensity to provide numerical estimates of fatalities in GPT-3.5.} In the same vein, we asked questions about the Turkish airstrikes such as \textit{``In the Turkish airstrike on August 8, 2015 in Midyat how many were killed?''}. 
%A complete list of all question prompts can be found in the Appendix.

\begin{figure}[htbp]
    \centering
    \includegraphics[width = 0.3\textwidth]{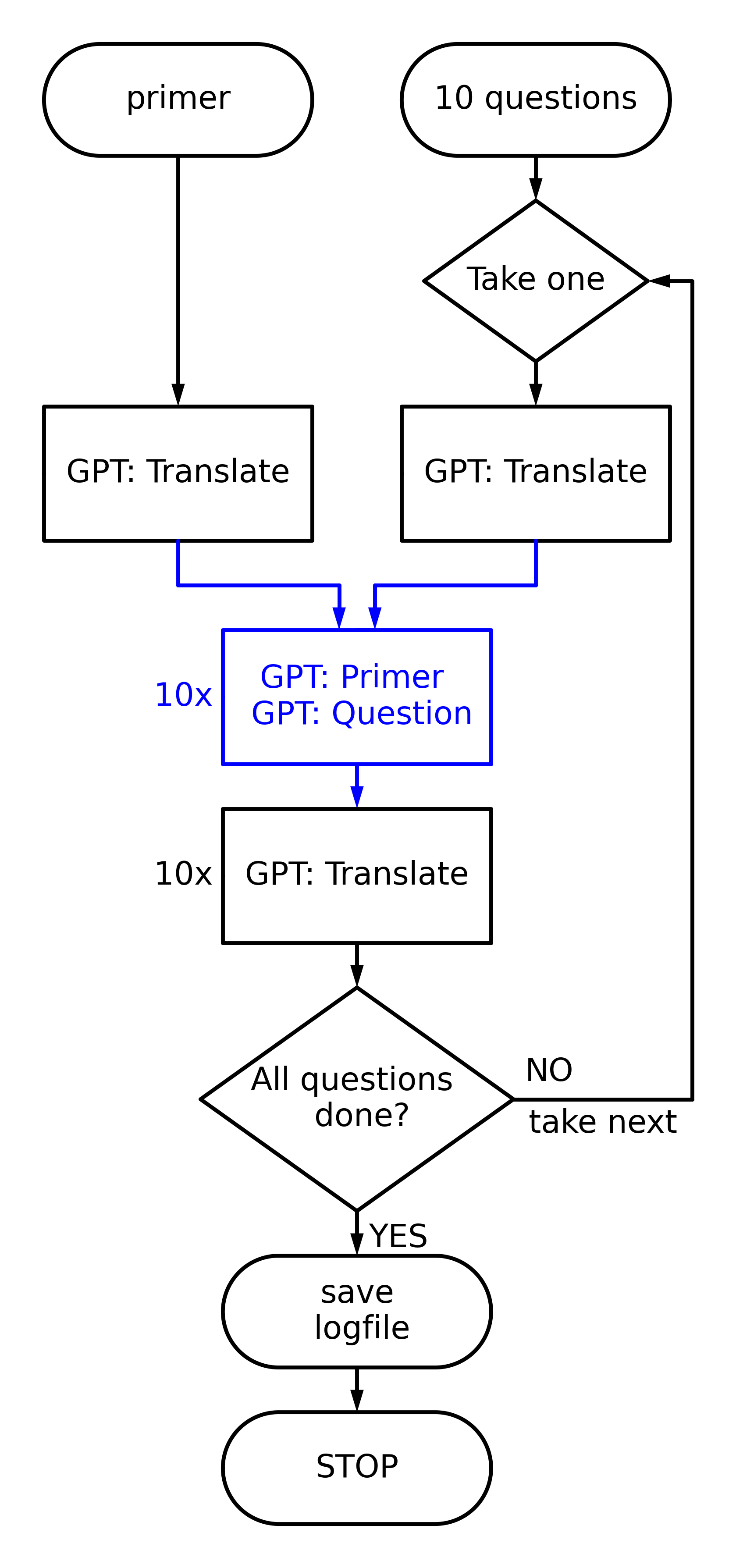}
    \caption[Query scheme]{\textbf{Query scheme:} The depicted scheme is used for all 4 languages. The square boxes denote GPT-3.5 requests. The blue boxes and lines denote communication being conducted completely in the target language.}
    \label{fig:scheme}
\end{figure}

To get a significant amount of data while keeping the coding effort acceptable our query procedure is fully automated. A short scheme of our approach is shown in figure \ref{fig:scheme}. We use Open-AI's Python API and the GPT-3.5-turbo algorithm, which is currently the cheapest and most widely used instance. For each query language, we follow the procedure described in in the figure and below, with each element consisting of a new instance to reduce memory effects and bias.  

Each of these instances consists of a system message (primer) defining the role of the instance, followed by the query itself. The number of maximum response tokens (each token is about one syllable, see \cite{open_ai_what_2023}) is always set to 1,000 in order to avoid unnecessarily verbose responses. 
% \footnote{Evidence shows that a lower temperature can enhance the performance of GPT-3.5 \citep[see]{peng_towards_2023}.}
The exact procedure for each language goes as follows:

\begin{enumerate}
    \item The primer is translated using the role: \textit{``You are a professional translator.''} and a temperature of zero to allow for reproducible translations. As a primer we use the phrase: \textit{``You are an expert of quantitative military history.''} This role provides fairly reproducible responses that involve exact numbers and are easy to code later on. We have tried standard phrases such as \textit{``You are a helpful assistant.''} or similar, but the amount of unusable answers involving non-exact quantities is too large and language-dependent. We assume that our assigned role gives similar results than longer chats with the API where someone just asks for exact numbers.
    \item We take a question and translate it using the same procedure as applied for the primer before.
    \item This is the main prompt. In an instance that only communicates in the query language we use the translated role and question to get a ``typical'' response in the user language. We do this 10 times in order to allow for statistical analysis. Setting the response `temperature' to 0.6 allows for a certain amount of randomness in the answers (for more detail see \cite{open_ai_chat_2023}).
    \item We automatically translate the answers back into English and save the whole conversation to allow for easy coding of the answers.
    \item We proceed with the next question.
    \item When all questions are completed, all queries and prompts are saved to a logfile.
    \item The coding and statistical analysis of the recorded responses is done manually in order to detect outliers, technical problems, etc. A random sample of questions and answers in each language were double-checked by native speakers to further detect any undesirable behavior/translation issues.
\end{enumerate}
All logfiles used for our analysis as well as a code sample of our query script are available online on Zenodo \citep[][]{kazenwadel_replication_2023}.\footnote{See: \href{https://doi.org/10.5281/zenodo.8181226}{https://doi.org/10.5281/zenodo.8181226}.}

% To get a more elaborate and longer text basis for classical text analysis such as word counts and sentiment analysis (see chapter TODO) we used a similar algorithm just with the primer: \textit{You are a helpful assistant.} and phrases such as: \textit{'What happened in the Israeli airstrike on February 7, 2006 in the Gaza strip? Tell me all the details about it.'}. This provided much more elaborate answers, thereby enabling quantitative text analysis.

\section{Evidence on numeric fatality estimates}

\begin{figure}[H]
    \centering
    \includegraphics[width = \textwidth]{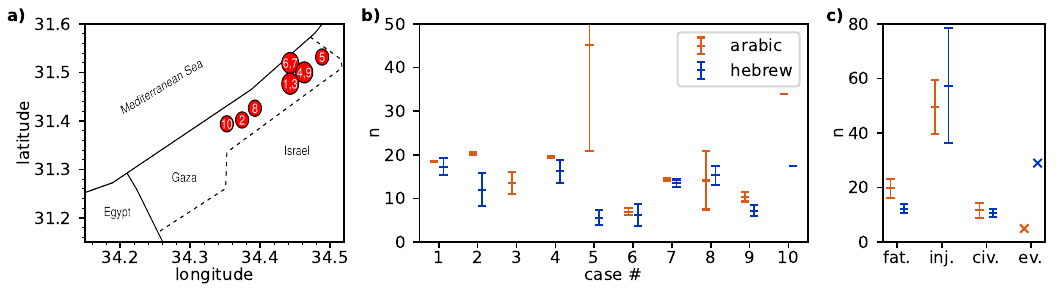}
    \caption[Quantitative results on the Arabic/Hebrew dyad]{\textbf{Quantitative results on the Arabic/Hebrew dyad:} a) Geographic distribution of airstrikes b) Number of recorded fatalities (military plus civilians) for each event. Blue are the fatalities recorded when asking in Hebrew, yellow the recorded fatalities when asking in Arabic. The errorbars denote the standard deviation of the mean, evasive answers were excluded. c) Number of total fatalities (fat.), injured (inj.), number of killed civilians averaged over all events (civ.). Evasive answers (ev.) are only observed when the answer says the event did not exist or it does not know of it. When the answer only says that GPT-3.5 does not know the exact fatalities the event is not coded as evasive, but still excluded from the statistical analysis.}
    \label{fig:arab_hebrew}
\end{figure}

\begin{figure}[H]
    \centering
    \includegraphics[width = \textwidth]{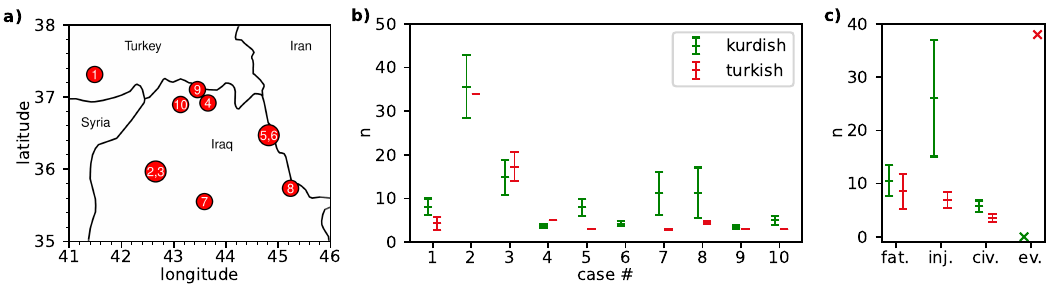}
    \caption[Quantitative results on the Kurdish/Turkish dyad]{\textbf{Quantitative results on the Kurdish/Turkish dyad:} a) Geographic distribution of airstrikes b) Number of recorded fatalities (military plus civilians) for each airstrike. Red are the fatalities recorded when asking in Turkish, green the recorded fatalities when asking in Kurdish. The errorbars denote the standard deviation of the mean. Evasive answers were excluded. c) Number of total fatalities (fat.), injured (inj.), number of killed civilians averaged over all events (civ.). Evasive answers are only observed when the answer says the event did not exist or it does not know of it (ev.). When the answer only says that GPT-3.5 does not know the exact fatalities the event is not coded as evasive. }
    \label{fig:kurman_turk}
\end{figure}

Our main results are presented in figure \ref{fig:arab_hebrew} for the Israeli-Palestinian conflict and in figure \ref{fig:kurman_turk} for the Turkish-Kurdish conflict. For each case, we provide the geolocated positions of the airstrikes in chronological order (panel a), GPT-3.5's quantitative estimates of the number of fatalities in each airstrike (panel b), and the fatality estimates averaged across all airstrikes (panel c). 

Beginning with the Israeli-Palestinian conflict, the evidence shows that the fatality estimates provided by GPT-3.5 are higher in Arabic than in Hebrew for 8 out of 10 airstrikes, while the pattern is reversed for one airstrike (case \#8). The error bars indicate the standard deviation and therefore the spread of the non-zero results in each case. This correlates with the temperature setting mentioned above and should therefore be interpreted with caution. On average, fatality estimates tend to be higher in GPT-3.5 responses in Arabic than in Hebrew, both for total fatalities and for civilian casualties.\footnote{Somewhat surprisingly, the average number of reported injured individuals is slightly higher in Hebrew compared to Arabic. This may reflect a tendency for Hebrew sources to report targeted individuals as injured rather than killed, resulting in a substitution process.} For one airstrike (case \#4), quantitative information on fatalities was only provided in Arabic. This discrepancy in the propensity to provide information about airstrikes is borne out in the aggregated analysis of evasive answers (ev.), which refer to cases where GPT-3.5 denied the airstrike in question or described another event. As shown in Figure \ref{fig:arab_hebrew} panel c), GPT-3.5 is more likely to respond that the respective airstrike did not occur when asked in Hebrew compared to questions in Arabic.

%mention the 13 guys in a cave in the turkish conflict
%We need a better name for Nans

Moving to the evidence for the Turkish-Kurdish conflict presented in figure \ref{fig:kurman_turk}, we find that queries to GPT-3.5 on Kurdish tend to result in higher fatality estimates compared to queries on Turkish. For 7 out of 10 airstrikes, the fatality numbers are higher when the information is searched in Kurdish compared to Turkish. Two airstrikes (cases \#3 \& \#4) deviate from this pattern, and for one airstrike fatality estimates are only provided in Kurdish (case \#6). Overall, the responses to GPT-3.5 queries on Kurdish, compared to Turkish, suggest that the airstrikes caused a higher human toll. On average, GPT-3.5 reports higher numbers of deaths, more civilians killed, and more people injured in Kurdish. Furthermore, GPT-3.5 is more likely to report that the airstrikes in question did not take place or that it is not aware of them when asked in Turkish. For example, there was a surprisingly high number of responses in the Turkish output reporting 13 dead Turkish soldiers in a cave. This is due to the abduction into a cave and subsequent execution of 13 Turkish citizens by the PKK in February 2021 \citep[][]{reuters_turkey_2021}. Notably, this case is frequently described in Turkish responses when GPT-3.5 is asked about Turkish airstrikes.

The discrepancy in evasive answers is especially striking for both conflicts. In the language of the attacker, we get a significant number of responses where GPT-3.5 states that it does not know of such an event (29 in Hebrew, 38 in Turkish). In the language of the targeted group, this behavior is less common (5 in Arabic, 0 in Kurdish). This is probably due to the fact that such events have a different news impact in the respective languages. When the number of media mentions of an event falls below a certain threshold (typically in the attacker's language), GPT-3.5 starts to mention other events or simply denies its existence. This could be even more problematic than the bias in the training data. Imagine a user that uses a typical search engine such as Google to search for the fatalities of an airstrike, she will get a small number of results, but the event will still be present at the top because the search engine cannot evade. When using Chat-GPT or similar language based models, the user will get an evasive answer resulting in an impression of zero fatalities.

% Our analysis reveals that ChatGPT provides a 27$\pm$11 percent lower fatality estimates when queried in the language of the attacker compared to the language of the targeted group.
 
 To sum up our quantitative results we calculated the percentage by which the estimate in the language of the attacker differs from the estimate in the language of the targeted group in each case. Cases where no casualties are reported or where all responses are evasive are excluded from the analysis. On average, reported casualties are 26 percent lower in Hebrew than in Arabic, with a standard deviation of nine percent. In the Turkish/Kurdish dyad we get a bias of 28$\pm$12 percent. Taking into account the evasive answers, this deviation would increase to more than 50 percent. This means that the reported numbers of casualties are significantly lower when asked in the language of the attacker.

\section{Evidence on word frequencies}

ChatGPT's tendency to characterize airstrikes as deadlier and bloodier in the language of the targeted group is reflected not only in the numerical estimates of fatalities, but also in the substantive information provided. To analyze the content of ChatGPT responses, we measured word frequencies in the logfiles of the GPT-3.5 output. 
Since we are not looking for precise death counts in this analysis, but for broader contextual information, we asked ChatGPT more broadly \textit{``what happened in the Israeli airstrike on date X in location Y?''} using the primer \textit{``You are a helpful advisor.''}.

\begin{figure}[H]
    \centering
    \includegraphics[width = \textwidth]{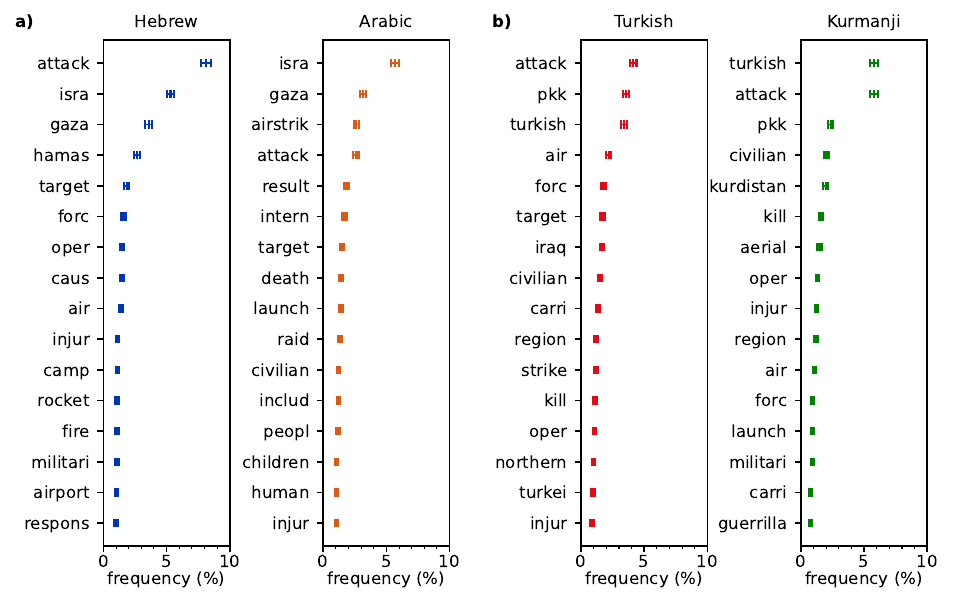}
    \caption[Word frequency analysis]{\textbf{Word frequency analysis:} Top 15 word stems in each language after removal of stopwords, with their respective word frequencies. The uncertainties are based on the shot noise limit a) Hebrew/Arabic dyad. b) Turkish/Kurdish dyad. }
    \label{fig:wordfreq}
\end{figure}

Figure \ref{fig:wordfreq} presents the most common, non-trivial words in the GPT-3.5 responses of each language.\footnote{We applied stemming and removed stopwords.} Some notable patterns emerge that support our claims of a language bias in GPT-3.5. The word frequency plot indicates that the stem ``hamas'' is the 4th most common word in Hebrew, while it is not among the top 15 terms in Arabic. In contrast, the terms ``civilian'' and ``children'' appear more frequently in Arabic. Similarly, the term ''pkk'' appears more frequently in the Turkish GPT-3.5 output. Tellingly, the stem ``kurdistan'' is the 4th most common term in Kurdish but does not appear among the top 15 terms in Turkish. Further, the terms ``civilians'' and ``guerilla'' have a greater relative frequency in Kurdish compared to Turkish.

To further explore the biases in the substantive information provided by GPT-3.5, we manually coded the relative frequency of claims of indiscriminate violence. Conflict research differentiates between selective and indiscriminate violence by asking whether the attacker uses force against the intended individuals while avoiding the use of force against those who were not targeted \citep[][]{gohdes_repression_2020,greitens_dictators_2016,kalyvas_logic_2006}. Violence is characterized as indiscriminate when it lacks precision, implying that `false positives' are among the victims. We conceptualize this important dimension of violence by manually coding statements about killed civilians and non-combatants such as children. As indiscriminate violence violates international humanitarian law, we further searched for references to the United Nations and human rights in the responses.

Civilian casualties are mentioned more than twice as often in the Arabic responses as in the Hebrew responses. Killed children are mentioned 6 times more often and female victims 3 times more often in the Arabic version. GPT-3.5’s responses in Arabic are also more likely to emphasize that these airstrikes violated international law. The United Nations is mentioned 13 times in the Arabic responses, while it is never mentioned in the Hebrew output. Moreover, the propensity differs by factor 11 that it was highlighted that these airstrikes were condemned by the international community. In contrast, the term ``terrorist’’ is mentioned more than 6 times more frequently in the Hebrew responses compared to the Arabic responses. %Overall, our analysis of word frequencies further highlights the language bias of GPT-3.5.
% killed children: 6 times more likely on Hebrew compared to Arabic
% 68% more frequently civilians mentioned on Arabic
% human rights more than 9 times more often on Arabic than Hebrew
% women three times more often on Arabic mentioned
% more than 11 times more likely to be mentioned that attack was condemened by international community
% 13 United Nations (none for Israel)
% more than twice as likely that killed civilians were mentioned
% terrorists more than 6 times as likely on Hebrew than Arabic

%Analogous to the Israeli-Palestinian conflict, we also searched for the frequencies of specific terms in the logfiles related to the Turkish-Kurdish conflict. We likewise find that terms related to victimization are more frequently mentioned in GPT-3.5 responses on the language of the targeted group. 
With regard to the Turkish-Kurdish conflict, we find that civilian casualties are mentioned about 50\% more often in Kurdish than in Turkish. Killed children appear 10 times more often in the Kurdish responses compared to the Turkish ones. Notably, the term ``innocent’’ appears only in the GPT-3.5 output in Kurdish. Furthermore, the term ``human rights’’ is mentioned 33\% more often in Kurdish than in Turkish. Overall, terms related to indiscriminate victimization or international condemnation appear more frequently in GPT-3.5 responses in Kurdish. In contrast, the term ``terrorist’’ is mentioned 8\% more often in the Turkish text compared to the Kurdish output.

% children more than 10 times as often on Kurdish
% 50% more likely that killed civilians were mentioned on Kurdish
% 33% more likely that human rights were mentioned on Kurdish 
% innocent only on Kurdish
% 8% more frequent the term terrorist on Turkish
\begin{comment}

\vspace{4mm}

Continue here with sentiment analysis
\begin{figure}[H]
    \centering
    \includegraphics[width = 0.5\textwidth]{sentimentbar.pdf}
    \caption[Word frequency analysis]{\textbf{Word frequency analysis:} Top 15 word stems in each language after removal of stopwords, with their respective word frequencies. The uncertainties are based on the shot noise limit a) Hebrew/Arabic dyad. b) Turkish/Kurdish dyad. }
    \label{fig:wordfreq}
\end{figure}
\end{comment}

\section{Discussion}

This study demonstrates that information on conflict-related violence generated by ChatGPT is affected by a substantial language bias. Using GPT-3.5, we show that the language model tends to produce higher fatality estimates when queried in the language of the targeted group compared to the language of the attacker. In the context of Turkish airstrikes against Kurdish targets, we find that GPT-3.5 produces higher fatality estimates if it is enquired in Kurdish compared to Turkish. In the same vein, GPT-3.5 reports higher fatality estimates in Arabic compared to Hebrew when asked about Israeli airstrikes in the Gaza Strip. Moreover, we show that airstrikes are described as more indiscriminate in the language of the targeted group, which is reflected in discrepancies in information about civilian casualties, killed children, and female victims. While it is well-established that language models can generate misinformation \citep{buchanan_truth_2021,solaiman_release_2019} and that they are linked to ethical and social risks \citep{bahrini_chatgpt_2023,weidinger_ethical_2021}, our study provides the first evidence of language bias in the context of conflict-related violence.

It is important to consider these findings in light of the underlying data generation process. By generating responses based on a multilingual corpus of online sources, ChatGPT's multilingual queries approximate large-scale language-specific media content analyses. Hence, although we specifically identify biases in the responses produced by the language model, they reproduce broader biases that are present in the training data. This suggests that to the extent that individuals consume online news in different languages, they tend to obtain different information about the extent and intensity of conflict-related violence.\footnote{Note that GPT-3.5 produces the modal response in the available online information. Of course, there is variation around this modal response.\nopagebreak} While this media bias is problematic in itself, ChatGPT makes it especially difficult for citizens to identify this bias. Critical consumers may be able to distinguish between high-quality and low-quality news sources, but they are less likely to understand the origins of the biases produced by the language model. Thus, by reproducing media biases under the guise of ostensible objectivity and `superhuman' intelligence, AI may exacerbate information discrepancies in armed conflicts. Furthermore, we find a novel source of bias that does not exist in classical search engines. By providing evasive answers in the language of the attacker, ChatGPT gives the impression that the airstrikes in question did not even occur. This amplifies the bias already present in the training data. %\citep{cheng_marked_2023,leavy_gender_2018,nadeem_gender_2020,lee_detecting_2018}.

This has far-reaching implications for multilingual armed conflicts. Public opinion plays a crucial role in armed conflicts as governments tend to rely on loyal troops and public support to wage war \citep{feinstein_rallyround_2022,tomz_public_2013,voeten_public_2006}. Misinformation may stir the public opinion against the opposing side and can even become a deadly self-fulfilling prophecy \citep{horowitz_deadly_2001}. Our findings suggest that citizens of states that have conducted airstrikes may underestimate the human toll of these attacks based on the information obtained through ChatGPT. In contrast, citizens of attacked groups may perceive these airstrikes as especially brutal and indiscriminate based on the information available in their language. These antithetical perceptions may contribute to radicalized identities and intensify dynamics of mutual blaming \citep[][]{hameleers_you_2022}. In so doing, information discrepancies may nurture grievances and ultimately reinforce conflicts within linguistic dyads.

Our findings have important implications beyond the specific context of armed conflict. It is plausible that similar language biases affect information generated by GPT-3.5 in other topic areas, especially where the training data is likewise heterogeneous and differs across languages. This is likely to be the case for other areas of contested information such as sensitive political issues, religious beliefs, or cultural identities. Future research could explore to what extent language biases in \mbox{GPT-3.5} responses are present in other topic areas, which languages are particularly susceptible to these biases, and if the same effects apply to other multilingual large language models as well. As the use of chatbots becomes more widespread, studies of their impact on public opinion are crucial. To conclude, our study provides a method for quantitatively analysing bias in large language models, offering a more robust quantitative alternative to classical sentiment analysis approaches. Using this new method, we show a significant bias between different user languages that could promote conflicts and information bubbles in the near future.

%%% TODO: We need a nice final paragraph %%%%%

%Additional empirical evidence from other contexts is required to evaluate the external validity of the findings presented in this study.

%Overall, our research offers novel empirical evidence on a language bias in ChatGPT and cautions users against using AI to obtain information on conflict-related violence. ChatGPT reproduces biases that are available in its training data, and the content of this training data differs systematically across languages. We currently live not only in times of an unprecedented availability of information sources, but that we also face a `collapse of the old news order' \citep[][p. 1868]{waisbord_truth_2018}. In these times of multilayered and complex news and communication environments it is highly important to develop a critical understanding of the information sources we use and to be aware of their inherent biases.

\newpage

\section*{Acknowledgements}
The authors would like to thank J. Holder and J. Weisser for many helpful comments and discussions. %, our native speakers K. Mahtouch, XXX  for the translation checks, as well as H. Zentner for providing their contacts. 
We acknowledge funding from evangelisches Studienwerk e.V. and from the International Postdoctoral Fellowship of the University of St. Gallen. We thank the `Konstanzer WG', especially P. Gebauer, S. Fonseca, L. Heyden, F. Böhringer, and H. Zentner for providing a highly conducive and stimulating environment for this research project. 
\newpage
\sloppy
\printbibliography

\newpage
\appendix
\renewcommand\thefigure{\thesection.\arabic{figure}} 
\setcounter{figure}{0}
\renewcommand\thetable{\thesection.\arabic{table}} 
\setcounter{table}{0}
\renewcommand{\thefootnote}{\arabic{footnote}a}
\setcounter{footnote}{0}
\setcounter{page}{1}
\renewcommand{\thepage}{A-\arabic{page}}

\begin{refsection}

%\section{Online Appendix}

\newpage
\printbibliography

\end{refsection}

\end{document}